\documentclass[letterpaper, 10 pt, conference]{ieeeconf}  

\IEEEoverridecommandlockouts                              

\usepackage{graphicx}
\usepackage{caption}
\usepackage{amsmath,amssymb,enumerate}

\usepackage{amsthm}
\usepackage{setspace}
\usepackage{booktabs}

\usepackage[usenames,dvipsnames,svgnames,table]{xcolor}
\usepackage[colorlinks=true,pdfpagemode=UseNone,citecolor=black,linkcolor=black,urlcolor=BrickRed,
pagebackref]{hyperref}

\usepackage{bbm}
\usepackage{mathtools}

\definecolor{note}{rgb}{0.1,0.1,1}
\definecolor{rephase}{rgb}{0.15,0.7,0.15}
\definecolor{bag}{rgb}{0.6,0.6,0.2}

\usepackage{algorithm}
\usepackage[noend]{algpseudocode}

\usepackage{gensymb}
\usepackage{xparse}

\usepackage{lipsum}

\usepackage{graphicx}
\usepackage{subcaption}

\usepackage{soul} 

\newtheorem{theorem}{Theorem}

\newtheorem{lemma}[theorem]{Lemma}

\def\realnumbers{\mathbb{R}}

\DeclareMathOperator*{\minimize}{minimize}

\renewcommand{\thefigure}{\arabic{figure}}

\newcommand{\Ccal}{\mathcal{C}}

\newcommand{\Fcal}{\mathcal{F}}
\newcommand{\Ical}{\mathcal{I}}
\newcommand{\Kcal}{\mathcal{K}}
\newcommand{\Lcal}{\mathcal{L}}

\newcommand{\Scal}{\mathcal{S}}
\newcommand{\Xcal}{\mathcal{X}}

\newcommand{\Zcal}{\mathcal{Z}}

\newcommand{\Exp}{\mathrm{Exp}}
\newcommand{\Log}{\mathrm{Log}}
\newcommand{\Jr}{\mathrm{J_r}}
\newcommand{\transpose}{\mathsf{T}}
\newcommand{\SO}{\mathrm{SO}}
\newcommand{\SE}{\mathrm{SE}}
\newcommand{\GL}{\mathrm{GL}}

\DeclareDocumentCommand{\X}{ O{} O{} }{\textbf{X}_{#1}^{#2}}
\DeclareDocumentCommand{\L}{ O{} O{} }{\textbf{L}_{#1}^{#2}}
\DeclareDocumentCommand{\S}{ O{} O{} }{\textbf{S}_{#1}^{#2}}
\DeclareDocumentCommand{\P}{ O{} O{} }{\textbf{P}_{#1}^{#2}}
\DeclareDocumentCommand{\Q}{ O{} O{} }{\textbf{Q}_{#1}^{#2}}
\DeclareDocumentCommand{\H}{ O{} O{} }{\textbf{H}_{#1}^{#2}}
\DeclareDocumentCommand{\J}{ O{} O{} }{\textbf{J}_{#1}^{#2}}
\DeclareDocumentCommand{\R}{ O{} O{} }{\textbf{R}_{#1}^{#2}}
\DeclareDocumentCommand{\A}{ O{} O{} }{\textbf{A}_{#1}^{#2}}
\DeclareDocumentCommand{\TH}{ O{} O{} }{\boldsymbol{\Theta}_{#1}^{#2}}
\DeclareDocumentCommand{\x}{ O{} O{} }{\textbf{x}_{#1}^{#2}}
\DeclareDocumentCommand{\e}{ O{} O{} }{\textbf{e}_{#1}^{#2}}
\DeclareDocumentCommand{\c}{ O{} O{} }{\textbf{c}_{#1}^{#2}}
\DeclareDocumentCommand{\C}{ O{} O{} }{\textbf{C}_{#1}^{#2}}
\DeclareDocumentCommand{\CM}{ O{} O{} }{\tilde{\textbf{C}}_{#1}^{#2}}
\DeclareDocumentCommand{\RM}{ O{} O{} }{\tilde{\textbf{R}}_{#1}^{#2}}
\DeclareDocumentCommand{\I}{ O{} O{} }{\textbf{I}_{#1}^{#2}}
\DeclareDocumentCommand{\O}{ O{} O{} }{\textbf{O}_{#1}^{#2}}
\DeclareDocumentCommand{\r}{ O{} O{} }{\textbf{r}_{#1}^{#2}}
\DeclareDocumentCommand{\t}{ O{} O{} }{\textbf{t}_{#1}^{#2}}
\DeclareDocumentCommand{\d}{ O{} O{} }{\textbf{d}_{#1}^{#2}}
\DeclareDocumentCommand{\b}{ O{} O{} }{\textbf{b}_{#1}^{#2}}
\DeclareDocumentCommand{\a}{ O{} O{} }{\textbf{a}_{#1}^{#2}}
\DeclareDocumentCommand{\dM}{ O{} O{} }{\tilde{\textbf{d}}_{#1}^{#2}}
\DeclareDocumentCommand{\p}{ O{} O{} }{\textbf{p}_{#1}^{#2}}
\DeclareDocumentCommand{\pM}{ O{} O{} }{\tilde{\textbf{p}}_{#1}^{#2}}
\DeclareDocumentCommand{\v}{ O{} O{} }{\textbf{v}_{#1}^{#2}}
\DeclareDocumentCommand{\w}{ O{} O{} }{\boldsymbol{\omega}_{#1}^{#2}}
\DeclareDocumentCommand{\wM}{ O{} O{} }{\tilde{\boldsymbol{\omega}}_{#1}^{#2}}
\DeclareDocumentCommand{\noise}{ O{} O{} }{\boldsymbol{\eta}_{#1}^{#2}}
\DeclareDocumentCommand{\FK}{ O{} }{\;\textit{\textbf{f}}_{#1}}
\DeclareDocumentCommand{\angleTheta}{ O{} O{} }{\boldsymbol{\theta}_{#1}^{#2}}
\DeclareDocumentCommand{\anglePhi}{ O{} O{} }{\boldsymbol{\phi}_{#1}^{#2}}
\DeclareDocumentCommand{\encoders}{ O{} O{} }{\boldsymbol{\alpha}_{#1}^{#2}}
\DeclareDocumentCommand{\encodersM}{ O{} O{} }{\tilde{\boldsymbol{\alpha}}_{#1}^{#2}}
\DeclareDocumentCommand{\offsets}{ O{} O{} }{\boldsymbol{\epsilon}_{#1}^{#2}}
\DeclareDocumentCommand{\Cov}{ O{} O{} }{\boldsymbol{\Sigma}_{#1}^{#2}}
\DeclareDocumentCommand{\Axis}{ O{} O{} }{\textnormal{Axis}_{#1}^{#2}}
\DeclareDocumentCommand{\using}{ O{} O{} }{\stackrel{\mathmakebox[\widthof{=}]{\text{eq.} (#1)}}{#2} \enspace}

\newcommand{\squeezeup}{\vspace{-3mm}}

\DeclareDocumentCommand{\zeros}{ O{} }{\textbf{0}_{#1}}

\title{\LARGE \bf
Legged Robot State-Estimation Through Combined Forward Kinematic and Preintegrated Contact Factors
}

\author{Ross Hartley, Josh Mangelson, Lu Gan, Maani Ghaffari Jadidi, Jeffrey M. Walls, \\ Ryan M. Eustice, and Jessy W. Grizzle
\thanks{The authors are with the College of Engineering, University of Michigan, Ann Arbor, MI 48109 USA {\tt\small \{{rosshart, mangelso, ganlu, maanigj, eustice, grizzle\}}@umich.edu}, with the exception of Jeff Walls who is with Toyota Research Institute.}%
}

\begin{document}
\maketitle
\begin{abstract} 
State-of-the-art robotic perception systems have achieved sufficiently good performance using Inertial Measurement Units (IMUs), cameras, and nonlinear optimization techniques, that they are now being deployed as technologies. However, many of these methods rely significantly on vision and often fail when visual tracking is lost due to lighting or scarcity of features. This paper presents a state-estimation technique for legged robots that takes into account the robot's kinematic model as well as its contact with the environment. We introduce forward kinematic factors and preintegrated contact factors into a factor graph framework that can be incrementally solved in real-time. The forward kinematic factor relates the robot's base pose to a contact frame through noisy encoder measurements. The preintegrated contact factor provides odometry measurements of this contact frame while accounting for possible foot slippage. Together, the two developed factors constrain the graph optimization problem allowing the robot's trajectory to be estimated. The paper evaluates the method using simulated and real sensory IMU and kinematic data from experiments with a Cassie-series robot designed by Agility Robotics. These preliminary experiments show that using the proposed method in addition to IMU decreases drift and improves localization accuracy, suggesting that its use can enable successful recovery from a loss of visual tracking.
\end{abstract}

\thispagestyle{empty}
\pagestyle{empty}

\section{Introduction and Related Work}
\label{sec:intro}

Legged locomotion enables robots to adaptively operate in unstructured and unknown environments with potentially rough and discontinuous ground~\cite{westervelt2007feedback}. The state-of-the-art control algorithms for dynamic biped locomotion are capable of providing stabilizing feedback for a biped robot blindly walking through sinusoidally varying ~\cite{da2017first} or discrete~\cite{nguyen2017dynamic} terrain; however, without perceiving the environment, the application of legged robots remains extremely limited. Accurate estimates of the robot's state and environment are essential prerequisites for both stable control and motion planning~\cite{bloesch2013state,fankhauser2014robot,bloesch2017state}. In addition, real-time performance of the state estimation and perception system is required to enable online decision making~\cite{fallon2014drift,nobiliheterogeneous}.

\begin{figure}[t!]
\vspace{.25cm}
	\centering
	\includegraphics[width=0.99\columnwidth]{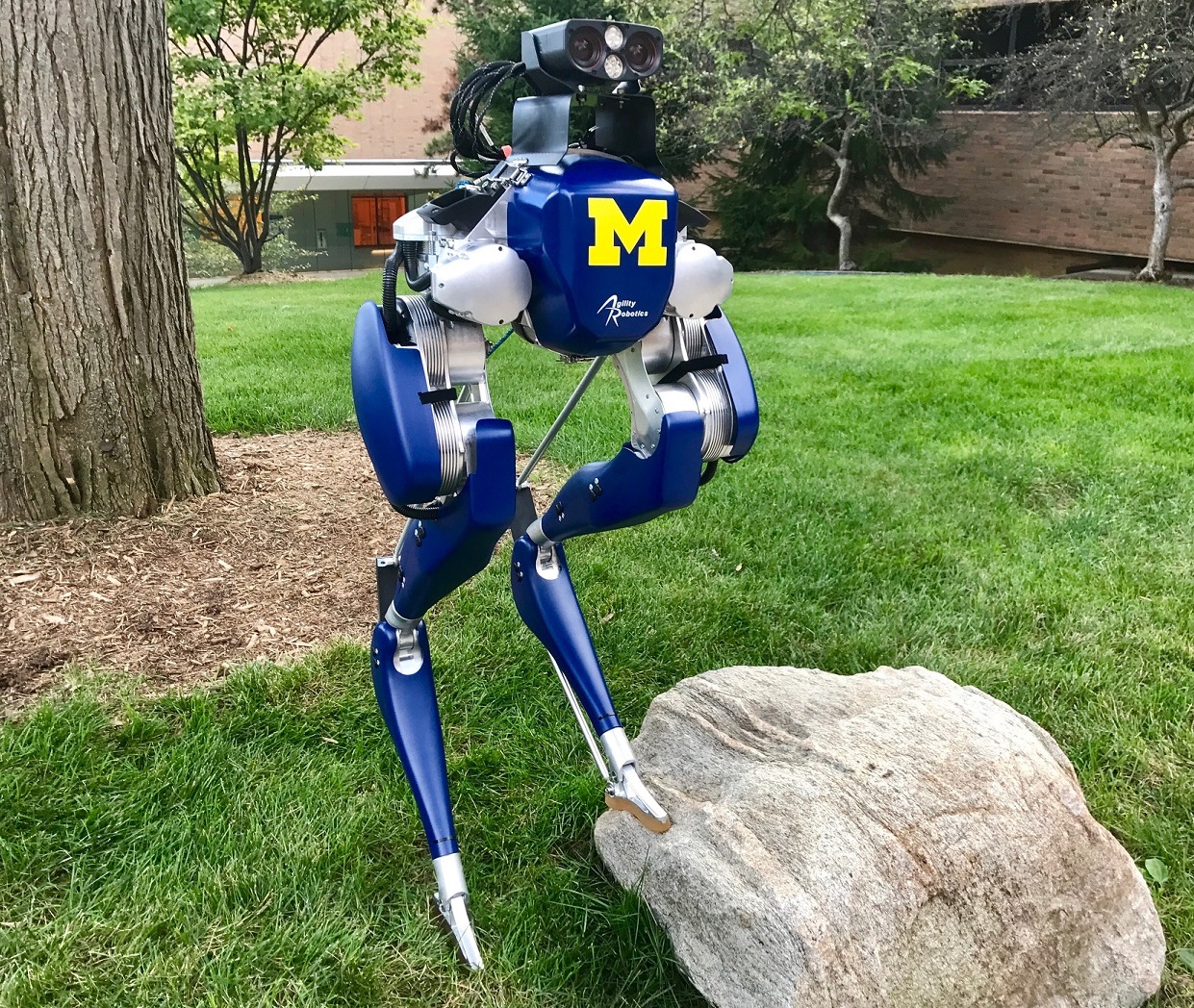}
	\caption{Experiments were conducted on a Cassie-series robot designed by Agility Robotics. The biped robot has 20 degrees of freedom, 10 actuators, joint encoders, an IMU, and a Multisense S7 stereo camera.}
	\label{fig:Motivation}
	\squeezeup\squeezeup
\end{figure}

Legged robots, unlike ground, flying, and underwater platforms, are in direct and switching contact with the environment. Leg odometry involves estimating relative transformations and velocity using kinematic and contact information, which can be noisy due to the encoder noise and foot slip~\cite{roston1991dead}. Typically, legged robots are equipped with additional sensors (IMUs, cameras, or LiDARs) which also provide independent, noisy odometry measurements (shown in Fig.~\ref{fig:Motivation}). Therefore, without a sound sensor fusion framework, the estimated trajectory can quickly become inaccurate as a consequence of significant drift over the traveled distance.

Filtering methods involve estimating the current state using all measurements up to the current time~\cite{anderson1979optimal}. Extended Kalman filters (EKFs) are often used to fuse high-frequency inertial and contact measurements to provide accurate velocity, and orientation estimates that are useful for the stabilizing feedback controller~\cite{bloesch2013state, rotella2014state}. However, the absolute position and yaw (rotation about gravity) have been shown to be unobservable~\cite{bloesch2013state}, which leads to the unbounded drift in these states. Re-observing landmarks over time using vision sensors allows for correcting the position and yaw estimates. However, landmark positions are unknown and have to be estimated alongside the robot's state. Over time, the accumulation of numerous landmarks can make the EKF computationally intractable for long-term state estimation and mapping~\cite{dissanayake2001solution}.

In contrast, smoothing methods estimate a discrete trajectory of states using all the available measurements \cite{anderson1979optimal}. Although the comparatively lower update rate may not be useful for the feedback controller, absolute position and yaw estimates can be corrected by relating the current pose to a previous one through loop closures~\cite{eustice2006exactly}. This allows for low-drift long-term state-estimation and mapping solutions. State-of-the-art visual-inertial odometry systems~\cite{forster2016manifold}, and, generally, Simultaneous Localization and Mapping (SLAM)~\cite{durrant2006simultaneous,cadena2016past}, use graphical models (factor graphs) and nonlinear optimization techniques to achieve a probabilistically sound sensor fusion framework and real-time performance by exploiting the sparse structure of the SLAM problem~\cite{thrun2004simultaneous,eustice2006exactly}. The high-dynamic motion and noise characterization is captured using preintegration of high-frequency sensors such as Inertial Measurement Units (IMUs)~\cite{lupton2012visual}. 

Although the factor graph framework has been successful, most methods heavily rely on visual information and are prone to failure when visual tracking is lost, often due to lighting or scarcity of features. In these scenarios, leg odometry is a way to reduce drift; and hereby, the incorporation of contact and encoder measurements into the factor graph framework are addressed. In this paper, we develop two novel factors that integrate the use of multi-link Forward Kinematics (FK) and the notion of contact between the robotic system and the environment into the factor graph smoothing framework. The forward kinematic factor relates a sensor frame (such as a  camera or IMU) to a contact frame through noisy encoder measurements. On the other hand, the contact factor preintegrates high-frequency foot contact measurements to describe the contact frame's movement over time. When combined, these novel factors constrain the robot's net movement, leading to improved state estimation. In particular, this work has the following contributions:

\begin{enumerate}[i.]
\item An FK factor that incorporates noisy encoder measurements to estimate an end-effector pose at any time-step;
\item rigid and point preintegrated contact factors that relate the contact frame pose between successive time-steps while accommodating noise from foot slip;
\item integration of leg odometry into the factor graph smoothing framework;
\item real-time implementation of the proposed FK and preintegrated contact measurement models on a Cassie-series biped robot. 
\end{enumerate}

Section~\ref{sec:prelim} provides the required preliminaries including the notation and mathematical prerequisites. We formulate the problem and our factor graph approach in Section~\ref{sec:problem}. Section~\ref{sec:fk} explains forward kinematic modeling. The forward kinematic factor is developed in Section~\ref{sec:ForwardKinematicsFactor}. Rigid and point contact factors are derived in Sections~\ref{sec:ContactFactor} and \ref{sec:point_contact_factor}, respectively. Simulation and experimental evaluations of the proposed methods on a 3D biped robot (Fig. \ref{fig:Motivation}) are presented in Section~\ref{sec:results}. Finally, Section~\ref{sec:conclusion} concludes the paper and provides future work suggestions.

\section{Preliminaries}
\label{sec:prelim}

In this section, we present preliminary materials necessary for the developments in the following sections. We first establish the mathematical notation where we assume readers are already familiar with basics of Lie groups, Lie Algebra, and optimization on matrix Lie groups~\cite{lee2015introduction,chirikjian2011stochastic,absil2009optimization}. Also, as this work is partly motivated by on-manifold IMU preintegration, we adopt the notation from~\cite{forster2016manifold}, where possible, so that readers can connect the two papers conveniently.

\subsection{Mathematical Notation and Background}

Matrices are capitalized in bold, such as in $\X$, and vectors are in lower case bold type, such as in $\x$. Vectors are column-wise and $[n]$ means the set of integers from $1$ to $n$, i.e.\@ \mbox{$\{1:n\}$}. The Euclidean norm is shown by $\lVert \cdot \rVert$. $\lVert \boldsymbol \e \rVert_{\Cov}^2 \triangleq \e[][\transpose] \Cov[][-1] \e$. The $n$-by-$n$ identity matrix and the $n$-by-$m$ matrix of zeros are denoted by $\I[n]$ and $\zeros[n,m]$ respectively. The vector constructed by stacking $x_i$, $\forall \ i \in [n]$ is denoted by $\mathrm{vec}(x_1,\dots,x_n)$. The covariance of a random vector is denoted by $\textnormal{Cov}(\cdot)$. Finally, we denote the base frame of the robot by $\textnormal{B}$, the world frame by $\textnormal{W}$, and contact frame by $\textnormal{C}$.

The \emph{general linear group} of degree $n$, denoted by \mbox{\small{$\GL_n(\realnumbers)$}}, is the set of $n\times n$ invertible matrices, where the group binary operation is the ordinary matrix multiplication. The \emph{special orthogonal group}, denoted by \mbox{\small{$\SO(3) = \{\R\in \GL_3(\realnumbers) | \R \R[][\transpose] = \I, \operatorname{det} \R = 1\}$}}, contains valid three-dimensional (3D) rotation matrices. The \emph{special Euclidean group}, denoted by \mbox{\small{$\SE(3) = \{ \mathbf{T} = \left[\begin{array}{cc} \R & \p \\ ~\mathbf{0}_3^\transpose & 1 \end{array} \right] \in \GL_4(\realnumbers) | \R \in \SO(3), \p \in \realnumbers^3 \}$}}, is the 3D rigid body motion group. Let \mbox{\small{$\w \triangleq \mathrm{vec}(\omega_1,\omega_2,\omega_3)$}}. The \emph{hat} operator is defined as $ \small{\w[][\wedge] \triangleq}
\scriptsize{\begin{bmatrix}
0 & -\omega_3 & \omega_2 \\
\omega_3 & 0 & - \omega_1 \\
-\omega_2 & \omega_1 & 0 \\
\end{bmatrix}}$, where the right hand side is known as the \emph{skew-symmetric matrix}. 

The Lie algebra of $\SO(3)$ is its tangent space at the identity together with Lie bracket and canonically determined as \mbox{\small{$\mathfrak{so}(3) = \{ \w[][\wedge] \in \GL_3(\realnumbers) | \exp(t \w[][\wedge]) \in \SO(3) \ \textnormal{and} \ t \in \realnumbers \}$}}. Conversely, the \emph{vee} operator maps a skew symmetric matrix to a vector in $\realnumbers^3$, that is $(\w[][\wedge])^{\vee} = \w$. In addition, $\forall \a,\b \in \realnumbers^3$ we have $\a[][\wedge] \b = -\b[][\wedge] \a$. The \emph{exponential map} associates an element of the Lie Algebra around a neighborhood of zero to a rotation matrix in $\SO(3)$ around a neighborhood of the identity and can be derived as \mbox{\small{$\exp(\boldsymbol{\phi}^\wedge) = \I + \dfrac{\sin(\lVert\boldsymbol{\phi}\rVert)}{\lVert\boldsymbol{\phi}\rVert} \boldsymbol{\phi}^\wedge + \dfrac{1-\cos(\lVert\boldsymbol{\phi}\rVert)}{\lVert\boldsymbol{\phi}\rVert^2}  (\boldsymbol{\phi}^\wedge)^2$}}; and using a first-order approximation reduces to \mbox{\small{$\exp(\boldsymbol{\phi}^\wedge) \approx \I + \boldsymbol{\phi}^\wedge$}}. For any $\lVert\boldsymbol{\phi}\rVert < \pi$, the \emph{logarithm map} uniquely associates a rotation matrix in $\SO(3)$ to an element of its Lie algebra $\mathfrak{so}(3)$; \mbox{\small{$\log(\R) = \dfrac{\varphi(\R-\R[][\transpose])}{2\sin(\varphi)}  \quad \textnormal{with} \quad \varphi = \cos^{-1}\left( \dfrac{\textnormal{tr}(\R) - 1}{2} \right)$}}. We use the following simplified notations from~\cite{forster2016manifold}:
\begin{equation} 
\small{
\label{eq:vectorized_maps}
\nonumber
\begin{array}{llll}
\Exp: \realnumbers^3 &\ni \anglePhi &\rightarrow \exp(\anglePhi[][\wedge])  &\in \SO(3) \\
\Log: \SO(3)         &\ni \R        &\rightarrow \log(\R)^\vee              &\in \realnumbers^3 .
\end{array}}
\end{equation}

The following first order approximations of above maps allow for relating increments in the Lie Algebra to increments in the Lie group and vice versa~\cite{forster2016manifold}
\begin{equation}
\small
 \label{eq:exp_map_approx}
 \Exp(\anglePhi + \delta \anglePhi) \approx \Exp(\anglePhi) \Exp(\Jr(\anglePhi) \delta \anglePhi)
\end{equation}
\begin{equation}
\small
 \label{eq:log_map_approx}
 \Log(\Exp(\anglePhi)\Exp(\delta \anglePhi)) \approx \anglePhi + \Jr^{-1}(\anglePhi) \delta \anglePhi
\end{equation}
where \mbox{\small{$\Jr(\anglePhi) = \I - \dfrac{1-\cos(\lVert \anglePhi \rVert)}{\lVert \anglePhi\rVert^2}\anglePhi[][\wedge] + \dfrac{\lVert \anglePhi \rVert - \sin(\lVert\anglePhi \rVert)}{\lVert \anglePhi \rVert^3}(\anglePhi[][\wedge])^2$}} is the right Jacobian of $\SO(3)$~\cite{chirikjian2011stochastic}. Finally, from the \emph{adjoint representation} of $\SO(3)$ we have
\begin{equation}
\small
 \label{eq:adjoint}
 \Exp(\anglePhi)\R = \R \Exp(\R[][\transpose] \anglePhi) .
\end{equation}

\subsection{Modeling Noise and Optimization on Matrix Lie Groups}
\label{sec:noise}

We model the uncertainty in $\SO(3)$ by defining a noise distribution in the tangent space and then mapping it to $\SO(3)$ via the exponential map~\cite{barfoot2017state, forster2016manifold}; $\tilde{\mathbf{R}} = \mathbf{R} \Exp (\epsilon)$ and $\epsilon \sim \mathcal{N}(\mathbf{0}, \mathbf{\Sigma})$, where $\R$ is a given noise-free rotation (the mean) and $\epsilon$ is a small
normally distributed perturbation with zero mean and covariance $\Cov$. Through approximating the normalization factor as constant, the negative log-likelihood of a rotation $\mathbf{R}$ given a measurement $\tilde{\mathbf{R}}$ distributed according to the defined perturbation is \mbox{\small{$\mathcal{L}(\R) \propto \frac{1}{2} \lVert \Log(\R[][-1] \tilde{\R}) \rVert^2_{\Cov} = \frac{1}{2} \lVert \Log (\tilde{\R}^{-1} \R) \rVert^2_{\Cov}$}} which is a \emph{bi-invariant Riemannian metric}. For the translation part of $\SE(3)$, noise can be characterized using the usual additive white Gaussian noise assumptions.

Solving an optimization problem where the cost function is defined on a manifold is slightly different from the usual problems in $\realnumbers^n$. The tangent vectors lie in the tangent space. As such, a \emph{retraction} that maps a vector in the tangent space to an element in the manifold is required~\cite{absil2009optimization}. Given a retraction mapping and the associated manifold, we can optimize over the manifold by iteratively \emph{lifting} the cost function of our optimization problem to the tangent space, solving the reparameterized problem, and then mapping the updated solution back to the manifold using the retraction. For $\SO(3)$, the exponential map is this retraction. For $\SE(3)$ we adopt the retraction used in \cite{forster2016manifold}:
\begin{equation}
\small
\mathcal{R}_T(\delta\anglePhi, \delta \p) = (\R \Exp ( \delta \anglePhi), \mathbf{p} + \R \delta \p), ~~\mathrm{vec}(\delta \anglePhi,\delta \p) \in \realnumbers^6 .
\end{equation}
\section{Problem Statement and Formulation}
\label{sec:problem}

In this section, we formulate the state estimation problem of the legged robot. The biped robot is equipped with a stereo camera, an IMU, joint encoders, and binary contact sensors on the feet. Without loss of generality, we assume the IMU and camera are collocated with the base frame of the robot. In current state-of-the-art visual-inertial navigation systems, the state includes the camera pose and velocity along with system calibration parameters such as IMU bias~\cite{forster2016manifold}. In the factor graph framework, independent measurements from additional sensors can be incorporated by introducing additional factors based on the associated measurement models. Foot slip is the major source of drift in leg odometry; as such, to isolate the noise at the contact point we augment the state at time-step $i$ to include the contact frame pose of both feet (in the world frame) \mbox{\small{$\C[i] \triangleq \{ \R[\textnormal{WC},l](i)\}_{l=1}^2$}} and \mbox{\small{$\d[i] \triangleq \{{}_\textnormal{W}\p[\textnormal{WC},l](i) \}_{l=1}^2$}}. However, without loss of generality, all following derivations are for a single contact frame.
Thus, the state at any time-step $i$ is represented as:
\begin{equation} \label{eq:state}
\small
\Scal_i \triangleq \{\R[i], \p[i], \v[i], \C[i], \d[i], \b[i]\}
\end{equation} 
where \mbox{\small{$\R[i] \triangleq \R[\textnormal{WB}](i)$}} is the base orientation, \mbox{\small{$\p[i] \triangleq {}_\textnormal{W}\p[\textnormal{WB}](i)$}} is the base position, \mbox{\small{$\v[i] \triangleq {}_\textnormal{W}\textbf{v}_\textnormal{B}(i)$}} is the base velocity, and \mbox{\small{$\b[i] \triangleq \b(i)$}} is the IMU bias. In addition \mbox{\small{$\Xcal_{k} \triangleq \bigcup_{i=1}^k \Scal_i$}} denotes the state up to time-step $k$.

Let $\Lcal_{ij} \in \SE(3)$ be a perceptual loop closure measurement relating poses at time-steps $i$ and $j$ ($j > i$) computed from an independent sensor, e.g.\@ using a point cloud matching algorithm. The forward kinematic measurements at time-step $i$ are denoted by $\Fcal_{i}$. The IMU and contact sensors provide measurements at higher frequencies. Between any two time-steps $i$ and $j$, we denote the set of all IMU and contact measurements by $\Ical_{ij}$ and $\Ccal_{ij}$, respectively. Let $\Kcal_k$ be the index set of time-steps (or key-frames) up to time-step $k$. We denote the set of all measurements up to time-step $k$ by $\Zcal_k \triangleq \{\Lcal_{ij}, \Ical_{ij},\Fcal_{i},\Ccal_{ij}\}_{i,j\in\Kcal_k}$.

\begin{figure}[t]
	\vspace{0.2cm}
	\centering 
	\includegraphics[width=0.99\columnwidth]{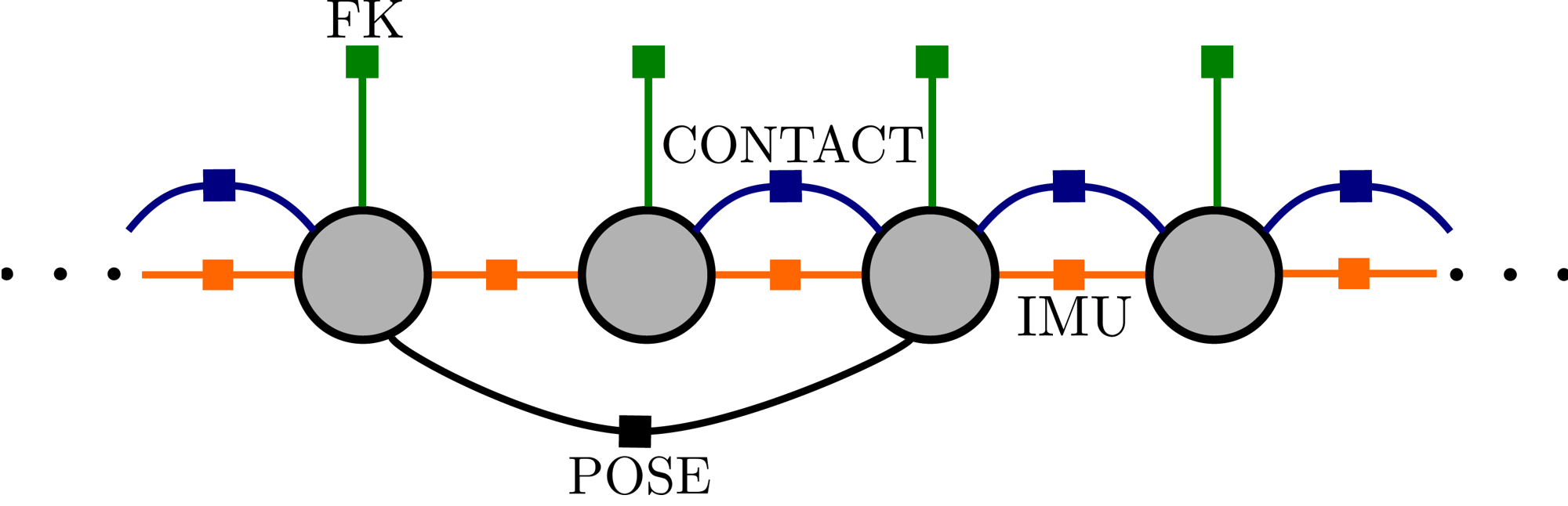}
	\caption{An example factor graph for the proposed system. Forward kinematic factors are added at each node and constrain the pose of the contact frames on the feet of the robot with respect to the robot base. Contact factors are added to the graph over time-steps where the given contact frame remained in contact with the environment. This framework enables the system to handle failures of the visual tracking or loop closure system (denoted here by general pose constraints).}
    \vspace{-0.7cm}
	\label{fig:factorgraph}
\end{figure}

By assuming the measurements are conditionally independent and are corrupted by additive zero mean white Gaussian noise, the posterior probability of the \emph{full SLAM} problem can be written as \mbox{\small{$p(\Xcal_{k}|\Zcal_k) \propto p(\Xcal_0) p(\Zcal_k|\Xcal_k)$}}, where
\begin{equation}
\small
  p(\Zcal_k|\Xcal_k) = \prod_{i,j\in \Kcal_k} p(\Lcal_{ij}|\Xcal_j) p(\Ical_{ij}|\Xcal_j) p(\Fcal_{i}|\Xcal_i) p(\Ccal_{ij}|\Xcal_j) .
\end{equation}

The \emph{Maximum-A-Posteriori} (MAP) estimate of $\Xcal_k$ can be computed by solving the following optimization problem:
\begin{equation}
  \small
  \label{eq:fullslam}
  \underset{\Xcal_k}{\minimize} \ -\log p(\Xcal_k|\Zcal_k)
\end{equation}
in which due to the noise assumption mentioned earlier is equivalent to the following nonlinear least-squares problem:
\begin{equation}
\small
\label{eq:map_nls}
  \begin{split}
  \underset{\Xcal_k}{\minimize} \ &\lVert \r[0] \rVert^2_{\Cov[0]} + \sum_{i,j\in\Kcal_k} \lVert \r[\Lcal_{ij}] \rVert^2_{\Cov[\Lcal_{ij}]} + \sum_{i,j\in\Kcal_k} \lVert \r[\Ical_{ij}] \rVert^2_{\Cov[\Ical_{ij}]}\\ 
  & + \sum_{i\in\Kcal_k} \lVert \r[\Fcal_{i}] \rVert^2_{\Cov[\Fcal_{i}]} + \sum_{i,j\in\Kcal_k} \lVert \r[\Ccal_{ij}] \rVert^2_{\Cov[\Ccal_{ij}]}
  \end{split}
\end{equation}
where $\r[0]$ and $\Cov[0]$ represents the prior over the initial state and serves to anchor the graph, $\r[\Lcal_{ij}]$, $\r[\Ical_{ij}]$, $\r[\Fcal_{i}]$, $\r[\Ccal_{ij}]$ are the residual terms associated with the loop closure, IMU, forward kinematic, and contact measurements respectively, i.e.\@ the error between the measured and predicted values given the state, and $\Cov[\Lcal_{ij}]$, $\Cov[\Ical_{ij}]$, $\Cov[\Fcal_{i}]$, $\Cov[\Ccal_{ij}]$ are the corresponding covariance matrices. A graphical example of this problem is shown in Fig.~\ref{fig:factorgraph}.

\section{Forward Kinematics}
\label{sec:fk}
Forward kinematics refers to the process of computing the position and orientation of an end-effector frame using measurements of the robot's joint positions. This section derives a representation of the forward kinematic functions needed to compute the pose of a contact frame relative to the base frame of the robot. 

\begin{figure} 
	\centering
    \vspace{1mm}
	\includegraphics[width=0.99\columnwidth]{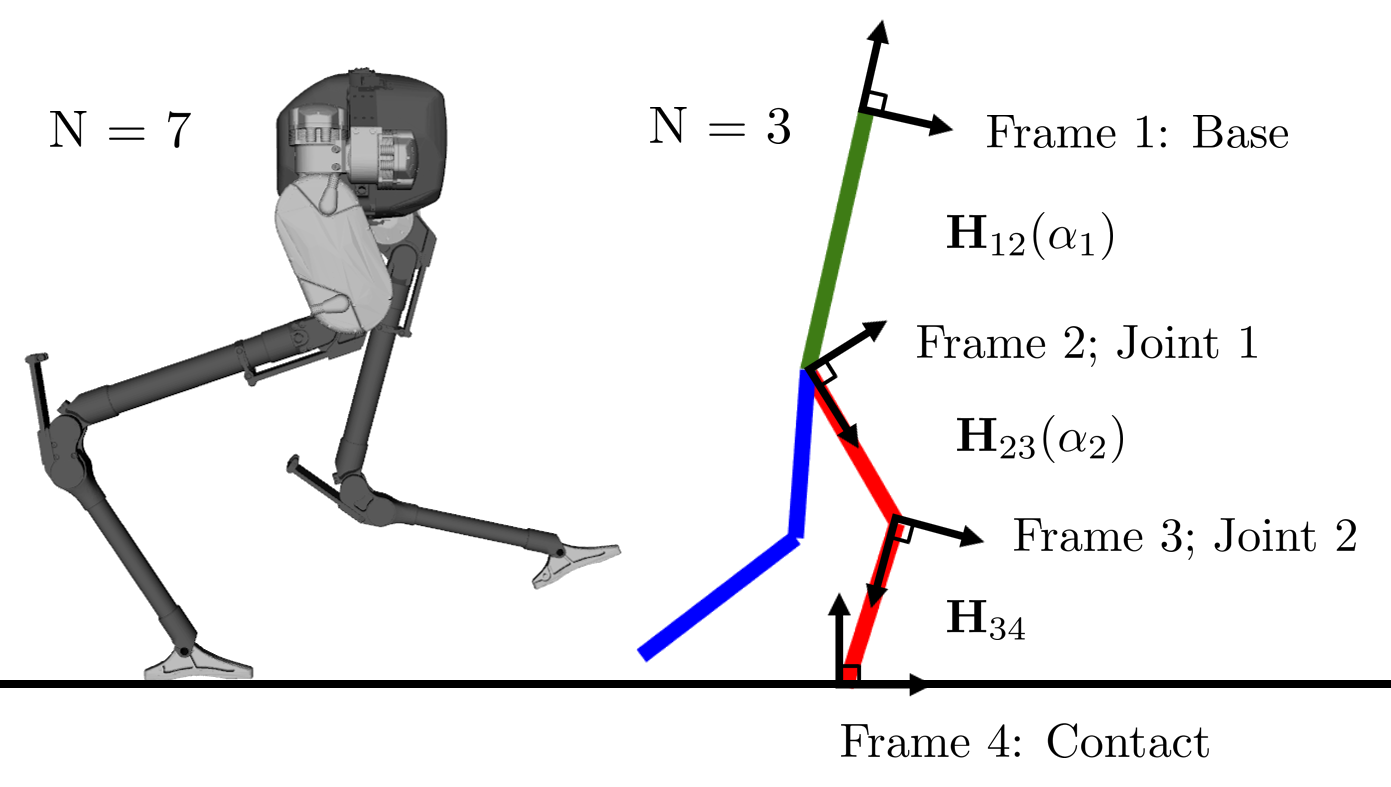}
	\caption{The contact frame is separated from the robot's base frame by $N$ links. Cassie (left) has 7 links between the base and the bottom of the toe. A simpler, planar biped (right) is shown to demonstrate the forward kinematics.}
	\label{fig:CoordinateFrames}
	\squeezeup\squeezeup
\end{figure}

Let the \emph{contact frame} be a coordinate frame on the robot located at a point of contact with the environment. This contact frame is separated from the robot's \emph{base frame}, by $N$ links. These $N$ links are assumed to be connected by $N-1$ revolute joints, each equipped with an encoder to measure the joint angle. The vector of encoder angles is denoted by $\encoders \in \realnumbers^{N-1}$. In total, this amounts to $N+1$ frames, where frames 1 and $N+1$ are the base and contact frame respectively, as shown in Fig. \ref{fig:CoordinateFrames}. The forward kinematics can be computed through a product of homogeneous transforms:
\begin{equation} 
\small
\label{eq:general_forward_kinematics}
\H(\encoders) = \prod_{n=1}^{N} \H[n,n+1](\alpha_n) \triangleq
\begin{bmatrix}
\FK[R](\encoders) & \FK[p](\encoders) \\
\zeros[1,3] & 1
\end{bmatrix} \in \SE(3)
\end{equation}
where $\FK[R](\encoders)$ and $\FK[p](\encoders)$ denote the rotation and position of the contact frame (relative to the base frame) as a function of encoder angles. The relative transformation between link frames \mbox{$n \in [N-1]$} and $n+1$ is given by:
\begin{equation}
\small
\H[n,n+1](\alpha_n) \triangleq
\begin{bmatrix}
\A[n]\Exp(\encoders[n][\dagger]) & \t[n] \\
\zeros[1,3] &     1
\end{bmatrix}
\end{equation}
where $\A[n] \in \SO(3)$ and $\t[n] \in \realnumbers^3$ are a constant rotation and translation defined by the kinematic model of the robot. Each joint is assumed to be revolute with an angle $\alpha_n \in \realnumbers$. The \emph{dagger} operator maps a scalar to a vector in $\realnumbers^3$ based on the joint's axis of rotation:
\begin{equation}
\small
\encoders[n][\dagger] \triangleq \mathrm{vec}(\alpha_n,0,0), \mathrm{vec}(0,\alpha_n,0), \ \textnormal{or} \ \mathrm{vec}(0,0,\alpha_n) .
\end{equation}
The final transformation between link frames $N$ and $N+1$ is constant and denoted by
\mbox{$\small{\H[N,N+1] \triangleq 
\begin{bmatrix}
\A[N] & \t[N] \\
\mathbf{0}_3^\transpose &     1
\end{bmatrix}}$}.
When \eqref{eq:general_forward_kinematics} is multiplied out, the orientation and position of the contact frame with respect to the base frame are:
\begin{equation}
\small
\label{eq:FK}
\FK[R](\encoders) = \A[1N+1](\encoders), ~~\text{and}~~ \FK[p](\encoders) = \sum_{n=1}^{N} \A[1n](\encoders)\t[n]
\end{equation}
where the relative rotation between link frames $i$ and $j$ is given by:
\begin{equation*}
\small
\A[ij](\encoders) \triangleq
\begin{cases}
\I  & \text{for } (i,j) \in [N],  i=j \\
\prod_{k=i}^{j-1} \A[k]\Exp(\encoders[k][\dagger])  & \text{for } (i,j) \in [N],  j > i \\
\A[iN] \A[N]  &\text{for } i \in [N-1], j=N+1 .\\
\end{cases}
\end{equation*}

%
%

Changes in joint angles affect the orientation of all link frames further down the kinematic tree. The following Lemma shows how the angle offsets propagate through the FK functions which is important for dealing with encoder noise. 
\begin{lemma}[Relative rotation between two frames~\cite{supplementary}]
Let $\boldsymbol{\beta} \in \realnumbers^N$ be a vector of joint angle offsets. The relative rotation between frames $i$ and $j$ can be factored into the nominal rotation and an offset rotation:
 \begin{equation} 
\small
\label{eq:joint_offset_propagation}
\A[ij](\encoders + \boldsymbol{\beta}) = \A[ij](\encoders) \prod_{k=i}^{j-1} \Exp\left( \A[k+1,j][\transpose](\encoders) \boldsymbol{\beta}^\dagger_k\right) 
\end{equation}
\end{lemma}


\section{Forward Kinematics Factor} 
\label{sec:ForwardKinematicsFactor}
The goal of this section is to formulate a general \emph{forward kinematic factor} that accounts for uncertainty in joint encoder readings. This factor can be included in a factor graph to allow estimation of end-effector poses. Here, the end-effector coincides with the contact frame. We note that this factor only constrains the relative transformation between the base and contact frames;
therefore, the full utility of this factor depends on a separate measurement of the contact frame described in Section~\ref{sec:ContactFactor}.


\subsection{Contact Pose through Encoder Measurements}
The encoder measurements are assumed to be affected by additive Gaussian noise, $\noise[][\alpha] \sim \mathcal{N}(0,\Cov[\alpha])$.
\begin{equation}
\small
\encodersM(t) = \encoders(t) + \noise[][\alpha](t)
\end{equation}

\noindent The orientation and position of the contact frame in the world frame are given by:
\begin{equation} 
\small
\label{eq:encoder_position_measurement}
\begin{alignedat}{2} 
\R[\textnormal{WC}](t) &= \R[\textnormal{WB}](t) \R[\textnormal{BC}](t) \\
{}_\textnormal{W}\p[\textnormal{C}](t) &= {}_\textnormal{W}\p[\textnormal{B}](t) + \R[\textnormal{WB}](t) _\textnormal{B}\p[\textnormal{C}](t) .
\end{alignedat}
\end{equation}

\noindent Rewriting \eqref{eq:encoder_position_measurement} in terms of the state \eqref{eq:state} and encoder measurements yields:
\begin{equation} 
\small
\label{eq:encoder_position_measurement_state}
\begin{alignedat}{2} 
\C(t) &= \R(t) \FK[R](\encodersM(t) - \noise[][\alpha](t)) \\
\d(t) &= \p(t) + \R(t) \FK[p](\encodersM(t)  - \noise[][\alpha](t)) .
\end{alignedat}
\end{equation}

The isolation of the noise terms in orientation and position of the contact frame are derived in the following Lemmas. The dependence on time is assumed, so $t$ is omitted for readability. 
\begin{lemma}[FK factor orientation noise isolation~\cite{supplementary}]
Using \eqref{eq:FK}, \eqref{eq:joint_offset_propagation}, and \eqref{eq:adjoint}, the rotation term and the noise quantity $\delta \FK[R]$ can be derived as:
\begin{equation} 
\small
\begin{alignedat}[b]{2} 
&\R[][\transpose] \C = \FK[R](\encodersM - \noise[][\alpha]) \\
&= \FK[R](\encodersM) \prod_{k=1}^{N-1} \Exp\left( -\A[k+1,N+1][\transpose](\encodersM) \noise[k][\alpha\dagger]\right) \triangleq \FK[R](\encodersM)\Exp(-\delta \FK[R])
\end{alignedat}
\end{equation}
\begin{equation} 
\small
\begin{alignedat}{2} 
\delta \FK[R] &= -\Log \left(\prod_{k=1}^{N-1} \Exp\left( -\A[k+1,N+1][\transpose](\encodersM) \noise[k][\alpha\dagger]\right)\right) \\
&\approx \sum_{k=1}^{N-1} \A[k+1,N+1][\transpose](\encodersM) \noise[k][\alpha\dagger]
\end{alignedat}
\end{equation}
Through repeated first order approximation, the noise quantity $\delta \FK[R]$ is approximately zero mean and Gaussian.
\end{lemma}
\begin{lemma}[FK factor position noise isolation~\cite{supplementary}]
Using \eqref{eq:FK}, \eqref{eq:joint_offset_propagation}, \eqref{eq:exp_map_approx}, and anticommutativity of skew-symmetric matrices, the position term can be approximated as:
\begin{equation} 
\small
\begin{alignedat}{2} 
\R[][\transpose]&(\d - \p) = \FK[p](\encodersM - \noise[][\alpha]) \\
&\approx \FK[p](\encodersM) + \sum_{k=1}^{N-1} \sum_{n=k}^{N-1} \A[1,n+1](\encodersM) \t[n+1][\wedge] \A[k+1,n+1][\transpose](\encodersM) \noise[k][\alpha\dagger] \\
&\triangleq \FK[p](\encodersM) - \delta \FK[p] 
\end{alignedat}
\end{equation}
The noise quantity $\delta \FK[p]$ is a linear combination of zero mean Gaussians, and is therefore also zero mean and Gaussian. 
\end{lemma}

Using these Lemmas, we can now write out the \emph{forward kinematic measurement model}:
\begin{equation} 
\small
\label{eq:FK_measurement_model}
\begin{alignedat}{2} 
\FK[R](\encodersM) &= \R[][\transpose] \C \, \Exp(\delta \FK[R]) \\
\FK[p](\encodersM) &= \R[][\transpose] (\d - \p) + \delta \FK[p]
\end{alignedat}
\end{equation}
where the forward kinematics noise characterized by $ \mathrm{vec}(\delta \FK[R], \delta \FK[p]) \sim \mathcal{N}(0,\Cov[\Fcal])$.

\subsection{Unary Forward Kinematic Factor}
The FK factor is a unary factor that relates the robot's base frame to an end-effector frame. Using \eqref{eq:FK_measurement_model}, we can write the residual errors, \mbox{$\small{\r[\Fcal_{i}] \triangleq \mathrm{vec}(\r[\FK[R_i]],\r[\FK[p_i]])}$} from \eqref{eq:map_nls}, at time $t_i$ as follows.
\begin{equation} 
\small
\begin{split}
\r[\FK[R_i]] &= \Log\left( \FK[R](\encodersM[i])^\transpose \R[i][\transpose] \C[i] \right)\\
\r[\FK[p_i]] &=  \R[i][\transpose] (\d[i] - \p[i]) - \FK[p](\encodersM[i])\\
\end{split}
\end{equation}
The forward kinematics noise can be rewritten as a linear system:
\begin{equation}
\small
\begin{bmatrix}
\delta \FK[R] \\
\delta \FK[p] \\
\end{bmatrix} = 
\begin{bmatrix}
\Q(\encodersM[i]) \\
\S(\encodersM[i]) \\
\end{bmatrix}
\noise[][\alpha\dagger]
\end{equation}
where $\noise[][\alpha\dagger] \triangleq \mathrm{vec}(\noise[1][\alpha\dagger], \noise[2][\alpha\dagger], \cdots, \noise[N-1][\alpha\dagger])$ and the columns of the $3 \times 3(N-1)$ matrices $\Q$ and $\S$ are given by:
\begin{equation}
\small
\begin{split}
\Q[i](\encodersM) &= \A[i+1,N+1][\transpose](\encodersM) \\
\S[i](\encodersM) &= -\sum_{n=i}^{N-1} \A[1,n+1](\encodersM) \t[n+1][\wedge] \A[i+1,n+1][\transpose](\encodersM) .\\
\end{split}
\end{equation}
The covariance can be computed using the linear noise model and the sensor covariance matrix $\Cov[\alpha\dagger]$ describing the encoder noise $\noise[][\alpha\dagger]$:
\begin{equation}
\small
\Cov[\Fcal_{i}] = 
\begin{bmatrix}
\Q(\encodersM[i]) \\
\S(\encodersM[i]) \\
\end{bmatrix}
\Cov[\alpha\dagger]
\begin{bmatrix}
\Q[][\transpose](\encodersM[i]) & \S[][\transpose](\encodersM[i])
\end{bmatrix} .
\end{equation}
The Jacobians for the forward kinematics factor are given in the supplementary material~\cite{supplementary}.

\section{Rigid Contact Factor} 
\label{sec:ContactFactor}
This section formulates a \emph{contact factor} based on the assumption that the contact frame remains fixed with respect to the world frame over time. Slip is accommodated by incorporating noise on the contact frames' velocities. When combined with the forward kinematic factor introduced in Section \ref{sec:ForwardKinematicsFactor}, an additional odometry measurement of the robot's base frame is obtained, which can improve the MAP estimate.

\subsection{Rigid Contact Model}
In addition to the encoders, it is assumed that a separate binary sensor can measure when the robot is in contact with the static world. If the contact is rigid (6-DOF constraint), then both the angular and linear velocity of the contact frame are zero; i.e ${}_\textnormal{C}\w[\textnormal{WC}](t) = {}_\textnormal{W}\v[\textnormal{C}](t) = \zeros[3,1]$. Therefore, 
\begin{equation} 
\small
\label{eq:contact_dynamics}
\begin{split}
\dot{\R}_\textnormal{WC}(t) &=  \R[\textnormal{WC}](t)({}_\textnormal{C}\w[\textnormal{WC}](t) + \noise[][\omega](t))^\wedge = \R[\textnormal{WC}](t) \noise[][\omega\wedge](t)\\
{}_\textnormal{W}\dot{\p}_\textnormal{C}(t) &= {}_\textnormal{W}\v[\textnormal{C}](t) + \R[\textnormal{WC}](t)\noise[][v](t) = \R[\textnormal{WC}](t) \noise[][v](t)
\end{split}
\end{equation}
where $\boldsymbol{\eta}^{\omega} \sim \mathcal{N}(\zeros[3,1],\Cov[\omega])$ and $\boldsymbol{\eta}^{v} \sim \mathcal{N}(\zeros[3,1],\Cov[v])$ are additive Gaussian noise terms that capture contact slip. This is similar to the EKF-based approach taken in \cite{bloesch2013state}. Both noise terms are represented in the contact frame and are rotated to align with the world frame. Rewriting \eqref{eq:contact_dynamics} in terms of the state vector yields:
\begin{equation} 
\small
\label{eq:contact_dynamics_state}
\begin{alignedat}{2}
\dot{\C}(t) &= \C(t) \noise[][\omega\wedge](t)\\
\dot{\d}(t)&= \C(t) \noise[][v](t) .
\end{alignedat}
\end{equation}

If the robot maintains rigid contact with the world from $t$ to $t + \Delta t$, Euler integration can be applied to obtain the pose of the contact frame at time $t + \Delta t$.
\begin{equation}
\small
\begin{split}
\C(t + \Delta t) &= \C(t) \Exp(\noise[][\omega d](t) \Delta t) \\
\d(t + \Delta t) &= \d(t) + \C(t) \noise[][v d](t) \Delta t
\end{split}
\end{equation}

\noindent Integrating from the initial time of contact, $t_i$, to the final time of contact, $t_j$, yields: 
\begin{equation} 
\small
\label{eq:contact_integration}
\begin{split}
\C[j] &= \C[i] \prod_{k=i}^{j-1} \Exp(\noise[k][\omega d] \Delta t) \\
\d[j] &= \d[i] + \sum_{k=i}^{j-1} \C[k] \noise[k][v d] \Delta t
\end{split}
\end{equation}
where $\noise[][\omega d]$ and $\noise[][v d]$ are discrete time noise terms computed using the sampling time; $ \textnormal{Cov}(\noise[][d](t)) = \dfrac{1}{\Delta t} \textnormal{Cov}(\noise(t))$.

\subsection{Rigid Contact preintegration}
We can now rearrange \eqref{eq:contact_integration} to create relative increments that are independent of the state at times $t_i$ and $t_j$.
\begin{equation} 
\small
\label{eq:contact_integration_relative}
\begin{split}
\Delta \C[ij] &= \C[i][\transpose] \C[j] = \prod_{k=i}^{j-1} \Exp(\noise[k][\omega d] \Delta t) \\
\Delta \d[ij] &= \C[i][\transpose](\d[j] - \d[i]) = \sum_{k=i}^{j-1} \Delta \C[ik] \noise[k][v d] \Delta t .
\end{split}
\end{equation}

Next, we wish to isolate the noise terms. First, we will deal with the rotation of the contact frame. The product of multiple incremental rotations can be expressed as one larger rotation. Therefore, 
\begin{equation} 
\small
\label{eq:contact_rotation_noise}
\begin{split}
\Delta \C[ij] &\triangleq \Delta \CM[ij] \Exp(-\delta \angleTheta[ij]) \\
\end{split}
\end{equation}
where $\Delta \CM[ij] = \I$ due to the rigid contact assumption. 
Furthermore, through repeated first order approximation, $\delta \angleTheta[ij]$ is approximately zero mean and Gaussian.
\begin{equation} 
\small
\delta \angleTheta[ij] = -\Log(\prod_{k=i}^{j-1}-\Exp(\noise[k][\omega d]\Delta t) ) \approx \sum_{k=i}^{j-1} \noise[k][\omega d] \Delta t
\end{equation}

Now, we can isolate the noise in the position of the contact frame by substituting \eqref{eq:contact_rotation_noise} into $\Delta \d[ij]$ and dropping the higher order noise terms. 
\begin{equation} 
\small
\label{eq:contact_position_noise}
\begin{split}
\Delta \d[ij] &\using[\ref{eq:exp_map_approx}][\approx]\sum_{k=i}^{j-1} \Delta \CM[ik](\I - \delta \angleTheta[ik][\wedge]) \noise[k][v d] \Delta t \approx -\sum_{k=i}^{j-1} \noise[k][v d] \Delta t \\
&\triangleq \Delta \dM[ij] - \delta \d[ij]
\end{split}
\end{equation}
where $\Delta \dM[ij] = 0$. The noise term, $\delta \d[ij] = \sum_{k=i}^{j-1} \noise[k][v d] \Delta t$, is zero mean and Gaussian. 

Finally, we arrive at the \emph{preintegrated contact measurement model}:
\begin{equation} 
\small
\begin{split}
\Delta \CM[ij] &= \C[i][\transpose] \C[j] \, \Exp(\delta \angleTheta[ij]) = \I[3] \\
\Delta \dM[ij] &= \C[i][\transpose] (\d[j] - \d[i]) + \delta \d[ij] = \zeros[3,1]
\end{split}
\end{equation}
where the rigid contact noise characterized by $ \mathrm{vec}(\delta \angleTheta[ij], \delta \d[ij]) \sim \mathcal{N}(0,\Cov[\Ccal_{ij}])$. More detailed derivations are provided in the supplementary material \cite{supplementary}.

\subsection{Preintegrated Rigid Contact Factor}

Once the noise terms are separated out, we can write down the residual errors, $\r[\Ccal_{ij}] = \mathrm{vec}(\r[\Delta \C[ij]],\r[\Delta \d[ij]])$  from \eqref{eq:map_nls}, as follows.
\begin{equation}
\small
\label{eq:contact_residuals}
\begin{split}
\r[\Delta \C[ij]] &= \Log( \C[i][\transpose] \C[j] ) \\
\r[\Delta \d[ij]] &= \C[i][\transpose] (\d[j] - \d[i]) 
\end{split}
\end{equation}

\noindent Furthermore, since both noise terms are simply additive Gaussians, the covariance can easily be computed. If the contact noise is constant, the covariance is simply:
\begin{equation} 
\small
\label{eq:contact_residuals_covariance}
\begin{split}
\Cov[\Ccal_{ij}]  = 
\begin{bmatrix}
\Cov[\omega] & \zeros[3,3] \\ 
\zeros[3,3] &  \Cov[v] \\
\end{bmatrix} \Delta t_{ij}
\end{split}
\end{equation}
where $\Cov[\omega]$ and $\Cov[v]$ are the continuous covariance matrices of the contact frame's angular and linear velocities, $\noise[][\omega]$ and $\noise[][v]$, and $\Delta t_{ij} = \sum_{k=i}^{j} \Delta t$.

If the contact noise is time-varying, the covariance can be computed iteratively. This would be particularly useful if the noise were modeled to depend on contact pressure. The iterative noise propagation and the Jacobians for the rigid contact factor are fully derived in the supplementary material~\cite{supplementary}.

\section{Point Contact Factor}
\label{sec:point_contact_factor}
The rigid contact factor can be modified to support additional contact types. For a point contact, the contact frame position remains fixed with respect to the world frame; however, the orientation can change over time. Therefore, $\Delta \CM[ij] \neq \I$, and subsequently becomes unobservable using only encoder and contact measurements. To track the contact noise appropriately, a gyroscope can be used. 

In the following section, the preintegrated contact factor is formulated as an extension to the preintegrated IMU factor described in~\cite{forster2016manifold}. This approach allows support for point contacts in our factor graph formulation. 

\subsection{Point Contact Model}
A point contact is defined as a 3-DOF constraint on the position of the contact frame. All rotational degrees of freedom are unconstrained. Since the relative orientation, $\Delta \CM[ij]$, is unobservable without the use of a gyroscope, we shall remove this term in the preintegrated contact factor by utilizing the preintegrated IMU measurements. Replacing $\C[k]$ in \eqref{eq:contact_integration} with its definition in \eqref{eq:encoder_position_measurement}, yields:
\begin{equation} 
\small
	\begin{split}
		\d[j] &= \d[i] + \sum_{k=i}^{j-1} \R[k] \FK[R](\encodersM[k] - \noise[k][\alpha]) \noise[k][v d] \Delta t .
	\end{split}
\end{equation}

We can rewrite this equation to be independent of the state at time $t_i$ and $t_j$.
\begin{equation} 
\small
	\begin{split}
		\R[i][\transpose](\d[j] - \d[i]) = \sum_{k=i}^{j-1} \Delta\R[ik] \FK[R](\encodersM[k] - \noise[k][\alpha]) \noise[k][v d] \Delta t
	\end{split}
\end{equation}
where $\Delta\R[ij] =  \R[i][\transpose] \R[j] = \prod_{i=1}^{j-1} \Exp\left( (\wM[k] - \b[k][g] - \noise[k][gd]) \Delta t \right)$ is the relative rotation increment from the IMU preintegration model \cite{forster2016manifold}.

The encoder measurements at time $t_i$ are already being used for the forward kinematic factor (Section \ref{sec:ForwardKinematicsFactor}). Therefore, to prevent information double counting, the first term in the summation can be replaced with the state estimate at time $t_i$. After a first-order approximation of the forward kinematics function, we arrive at the \textit{preintegrated point contact position measurement} $\Delta \dM[ij]$ and its noise $\delta \d[ij]$:
\begin{equation}
\small
	\begin{split}
		\R[i][\transpose](\d[j] - \d[i]) 
		& \approx \R[i][\transpose]\C[i]\noise[i][vd] \Delta t + \sum_{k=i+1}^{j-1} \Delta\RM[ik] \FK[R](\encodersM[k]) \noise[k][v d] \Delta t  \\
		&\triangleq \Delta \dM[ij] - \delta \d[ij]
	\end{split}
\end{equation}
where, again, $\Delta \dM[ij] = \zeros[3,1]$. The full derivation is detailed in the supplementary material~\cite{supplementary}.

Up to a first order approximation, $\delta \d[ij]$ is zero mean, Gaussian, and does not depend on the encoder or IMU noise, i.e.\@ it is decoupled~\cite{supplementary}. Since the contact measurement is uncorrelated to the IMU / forward kinematics measurements and all are jointly Gaussian, the measurements are all independent. Therefore, the point contact factor can be written separately to the IMU factor.

\subsection{Preintegrated Point Contact Factor}
Once the noise terms are separated out, we can write down the residual error:
\begin{equation} 
\small
\label{eq:point_contact_residual}
\r[\Ccal_{ij}] = \r[\Delta \d[ij]] = \R[i][\transpose] (\d[j] - \d[i]) . 
\end{equation}
The noise propagation can be written in an iterative form:
\begin{equation} 
\small
\label{eq:iterative_noise_point_contact}
	\delta \d[ik+1] =
\begin{cases}
    \delta \d[ik] - \R[k][\transpose] \C[k] \noise[k][vd] \Delta t & \text{for  } k = i\\
	\delta \d[ik] - \Delta \RM[ik] \FK[R](\encodersM[k]) \noise[k][v d] \Delta t & \text{for  } k > i\\
\end{cases}
\end{equation}
which allows us to write the covariance propagation as a linear system (starting with $\Cov[\Ccal_{ii}] = \zeros[3,3]$):
\begin{equation}
\small
	\Cov[\Ccal_{ik+1}]  = \Cov[\Ccal_{ik}] + \textbf{B} \Cov[vd]  \textbf{B}^\transpose
\end{equation}
where,
\begin{equation}
\small
	\textbf{B}  = 
    \begin{cases}
         \R[k][\transpose] \C[k] \Delta t & ~~~~ \text{for  } k = i \\
         \Delta\RM[ik] \FK[R](\encodersM[k]) \Delta t  & ~~~~ \text{for  } k > i .\\
    \end{cases}
\end{equation}
The covariance of the discrete contact velocity noise $\noise[k][vd]$ is denoted by $\Cov[vd]$. These equations are derived in complete detail in the supplementary material \cite{supplementary}.

\section{Simulations and Experimental Results}
\label{sec:results}
In this section, we evaluate the proposed method and factors. We implemented both factors in GTSAM~\cite{dellaert2012factor} using
iSAM2 as the solver~\cite{kaess2012isam2}. To handle IMU preintegration we used the implementation built into GTSAM~4~\cite{forster2016manifold, carlone2014eliminating}. Both simulation and real-world experiments used the Cassie-series robot developed by Agility Robotics (shown in Fig.~\ref{fig:Motivation}). 

\subsection{Simulated Evaluation using SimMechanics}

For initial evaluation, we used SimMechanics to simulate a full model of Cassie walking along a curved path. The simulator provided true acceleration, angular velocity, joint angle, and contact values as well as ground-truth trajectory position and velocity. 

We generated IMU measurements by adding Gaussian noise and bias according to the models described in \cite{forster2016manifold}. We also simulated loop closure (LC) measurements using the ground-truth trajectory and corrupted them using the methods detailed in Section \ref{sec:noise}. These (local) loop closure factors were added to every other node in the graph in an attempt to simulate the results of visual odometry or scan matching algorithms. Finally, we added white Gaussian noise to the contact and joint angle values to generate forward kinematic and contact measurements. Table \ref{tab:simParams} shows the noise parameters used. A new node in the graph was added every time contact was made or broken with the environment (approximately 3 per second).

\begin{figure}[t!]
\vspace{0.3cm}%
\centering%
\includegraphics[width=0.99\columnwidth]{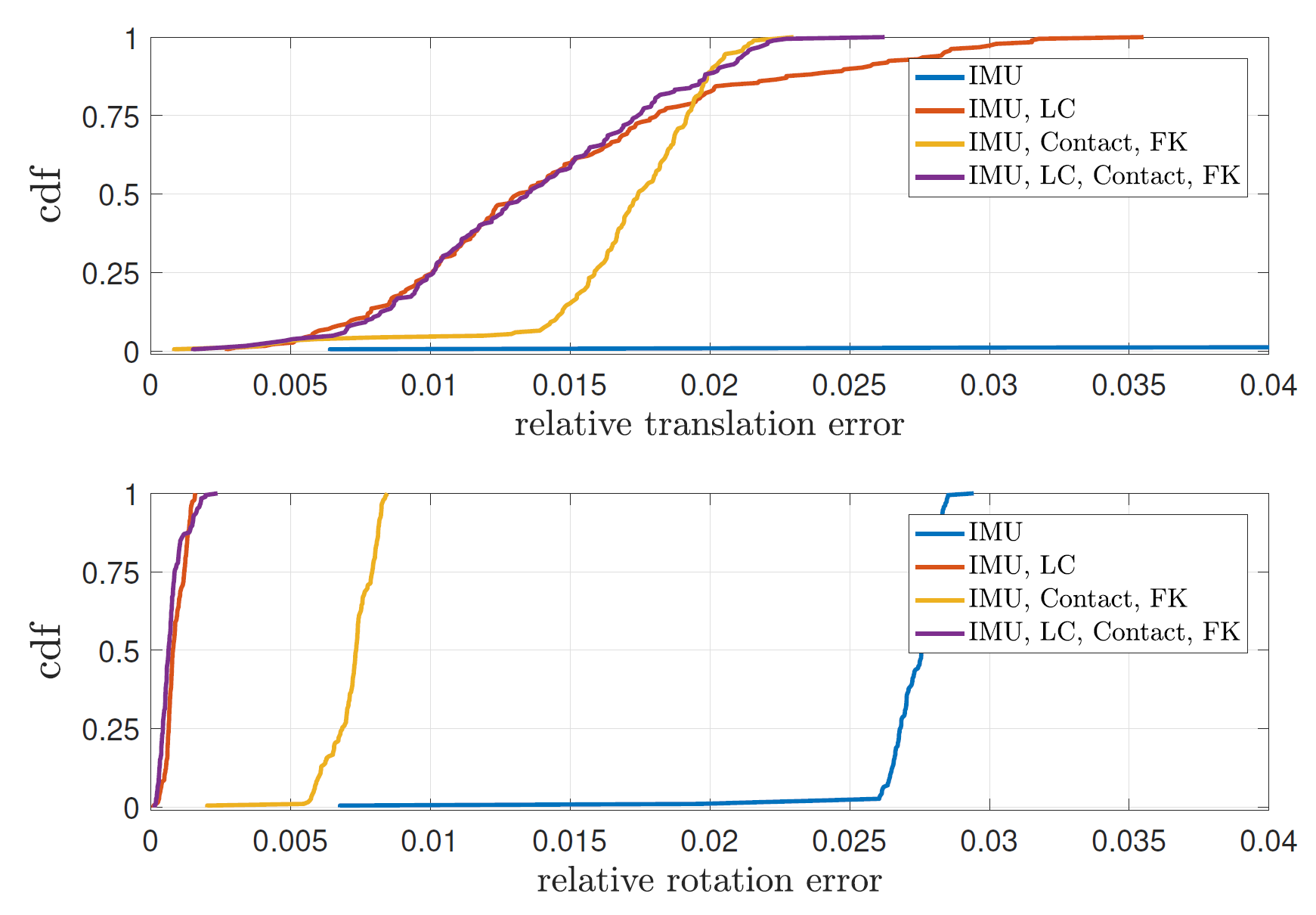}%
\caption{The cumulative distribution function of the norm of the translation and rotation errors for consecutive poses generated using a model of Cassie in SimMechanics. Measuring error in this way allows us to evaluate the drift of the different combinations of factors. Forward kinematics, contact, and IMU significantly outperform IMU and even outperforms the local loop closure (LC) and IMU in some cases.}%
  	\label{fig:sim_traj_error}%
\squeezeup
\end{figure}

\begin{table}[th!]
\footnotesize
	\centering
	\caption{Simulation Noise Parameters}
	\label{tab:simParams}
     \resizebox{0.5\columnwidth}{!}{%
	{\renewcommand{\arraystretch}{1.0}%
	\begin{tabular}{l|l}
            \toprule
			Noise			        & st. dev.\\
            \midrule
			Acceleration			& $0.0307 ~\text{m/s}^2$\\
			Angular Velocity        & $0.0014 ~\text{rad/s}$\\
			Accelerometer Bias      & $0.005 ~\text{m/s}^2$\\
			Gyroscope Bias          & $0.0005 ~\text{rad/s}$\\
			Loop Closure Translation  & $0.1 ~\text{m}$  \\
			Loop Closure Rotation & $0.0873 ~ \text{rad}$  \\
			Contact Linear Velocity  & $0.1 ~\text{m/s}$  \\
			Joint Encoders  & $0.00873 ~\text{rad}$  \\
            \bottomrule
	\end{tabular}}}
	\squeezeup
\end{table}

In this experiment, we compared the following combinations of factors: IMU, LC and IMU, IMU and Contact/FK, LC and IMU and Contact/FK. 
Fig. \ref{fig:sim_traj_error} shows the cumulative distribution function of the norm of the translation and rotation errors computed by comparing the relative pose between consecutive time-steps of the trajectory estimated by each combination of methods to the ground-truth. This metric allows for evaluating the drift of the different combinations of factors. The translational error is computed in the usual manner. The rotation error is computed using $\lVert \Log(\R[\mathrm{true}][\top] \R[\mathrm{est}])\rVert$. 

Forward kinematic, contact and IMU combined, significantly outperform IMU alone. This is partly because the contact and forward kinematic factors constrain the graph enabling the estimator to
solve for the IMU bias. Also, we note that in translation about $20\%$ of the time the combination of our proposed factors and the IMU outperforms the combination of the camera and the IMU. Finally, 
we note that the combination of all factors is the most successful.

These results suggest that the proposed method can be used to increase the overall localization accuracy of the system as well as to handle drop-outs in visual tracking. 

\subsection{Real-world}
We evaluated our factor graph implementation using real measurement data collected from a Cassie-series robot. This data included IMU measurements and joint encoders values. Cassie has two springs, located on each leg, that are compressed when the robot is standing on the ground. The binary contact measurement was computed using measurements of these spring deflections. The data was collected at 2KHz.

\begin{figure}[t] 
\vspace{0.3cm}%
	\centering 
    \includegraphics[width=0.98\columnwidth]{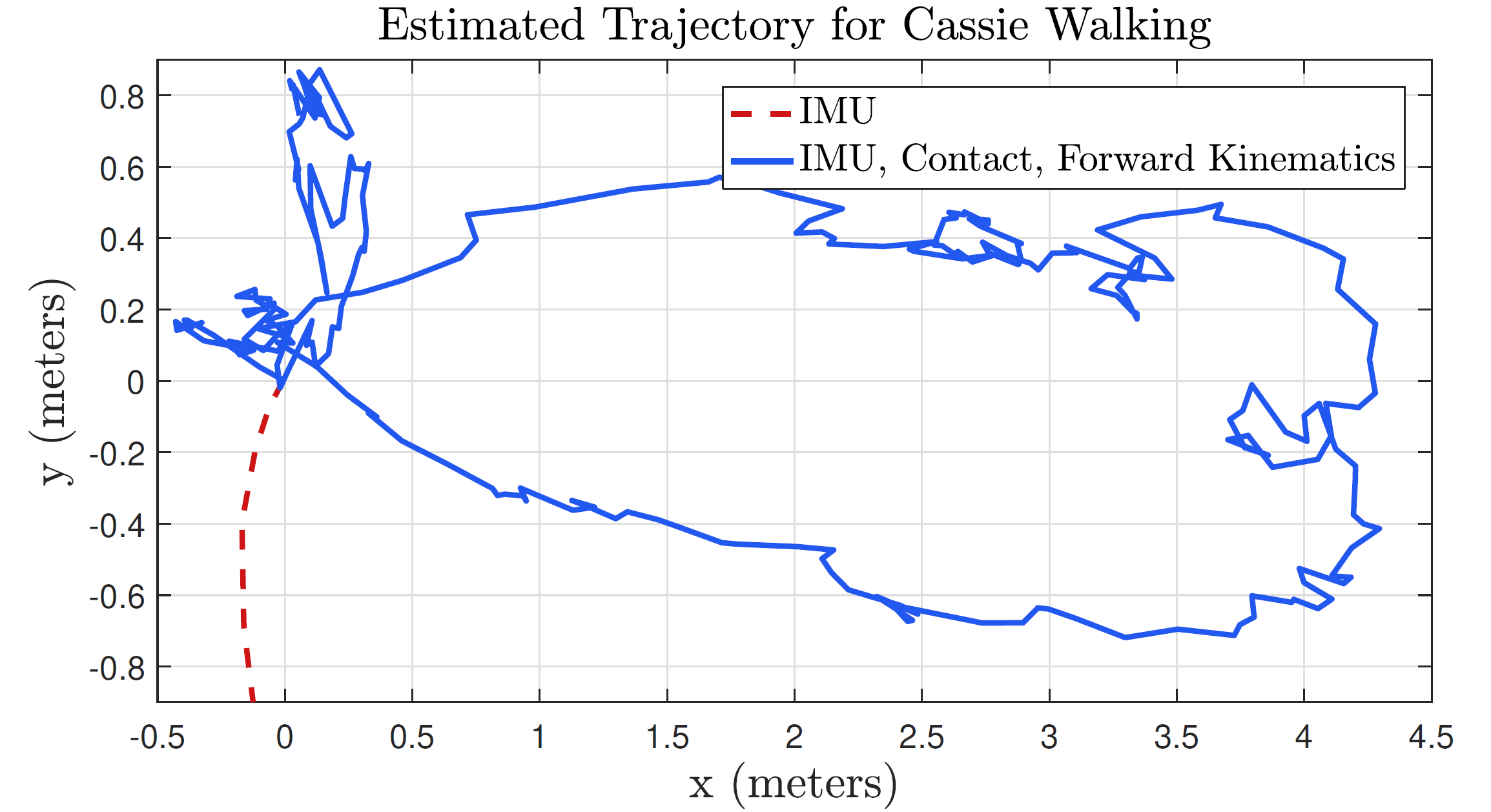}
    \caption{Estimated trajectory of Cassie experiment data using IMU, forward kinematic, and contact factors. The robot walked in a loop around the lab, starting and ending at approximately the same pose. The video of the experiment is shown at \url{https://youtu.be/QnFoMR47OBI}. Compared to the IMU only estimation (dashed red line), the addition of contact and forward kinematic factors significantly improves the state estimate. The end-to-end translation error was approximately $1.3~m$.}
 \label{fig:ExperimentResults}
 \squeezeup
\end{figure}

Using a controller provided by Agility Robotics, we walked Cassie for about 100 seconds in a loop around a 4.5-meter section of our lab, starting and ending in approximately the same location. Figure \ref{fig:ExperimentResults} compares the estimated trajectory computed using IMU, contact and forward kinematic factors with using only IMU factors. Due to noisy measurements, integration errors, and IMU bias, the IMU only estimate quickly drifts far away from the actual trajectory. After including the contact and kinematic factors, the state estimate is significantly improved resulting in a final translation error of approximately $1.3~\text{m}$.

Although the improved performance over IMU alone may appear trivial, this preliminary experiment validates the primary motivation of this work. Our novel forward kinematic and contact factors (in addition to IMU) can be used to improve odometry in factor graphs when vision systems fail. In this experiment, no visual odometry or loop closures were added, yet the addition of leg odometry enables the estimation of biases and reduces drift. In the future, we are working to incorporate visual odometry and loop closure constraints from the attached Multisense S7 camera (developed by Carnegie Robotics). We also plan to extensively test these new factors over a range of walking gaits and terrains in addition to collecting motion capture data to provide a proxy for ground truth.

\section{Conclusion}
\label{sec:conclusion}
We developed two novel factors within the factor graph framework to provide reliable leg odometry for bipedal locomotion. In particular, the forward kinematic factor computes measurements in the form of relative 3D transformation between the base and contact frames, while the preintegrated contact factor incorporates high-frequency contact measurements as constraints between any two successive contact frames in the graph. The evaluation showed that in the absence of accurate sensory data such as those from a stereo camera, the probabilistic fusion of IMU, FK, and contact factors enable the robot to track its trajectory while estimating IMU biases.

We are currently in the process of developing the open source implementation of the proposed techniques and performing additional evaluations in real experiments. In the future, we plan to use these techniques to build up both local and global maps that can be used for path planning as well as to inform the robot controller about the surrounding terrain.

\section*{ACKNOWLEDGMENT}
\footnotesize{
The authors thank Agility Robotics for both designing the robot and the walking controller used for the reported experiments. Funding for R. Hartley, L. Gan, and M. Ghaffari Jadidi is given by the Toyota Research Institute (TRI), partly under award number N021515, however this article solely reflects the opinions and conclusions of its authors and not TRI or any other Toyota entity. Funding for J. Mangelson is provided by the Office of Naval Research (ONR) under award N00014-16-1-2102. Funding for J. Grizzle was in part provided by TRI and in part by NSF Award No.~1525006.}
\vspace{-2.5mm}

\bibliographystyle{bib/IEEEtran}
\bibliography{bib/references}

\begin{thebibliography}{10}
\providecommand{\url}[1]{#1}
\csname url@rmstyle\endcsname
\providecommand{\newblock}{\relax}
\providecommand{\bibinfo}[2]{#2}
\providecommand\BIBentrySTDinterwordspacing{\spaceskip=0pt\relax}
\providecommand\BIBentryALTinterwordstretchfactor{4}
\providecommand\BIBentryALTinterwordspacing{\spaceskip=\fontdimen2\font plus
\BIBentryALTinterwordstretchfactor\fontdimen3\font minus
  \fontdimen4\font\relax}
\providecommand\BIBforeignlanguage[2]{{%
\expandafter\ifx\csname l@#1\endcsname\relax
\typeout{** WARNING: IEEEtran.bst: No hyphenation pattern has been}%
\typeout{** loaded for the language `#1'. Using the pattern for}%
\typeout{** the default language instead.}%
\else
\language=\csname l@#1\endcsname
\fi
#2}}

\bibitem{westervelt2007feedback}
E.~R. Westervelt, J.~W. Grizzle, C.~Chevallereau, J.~H. Choi, and B.~Morris,
  \emph{Feedback control of dynamic bipedal robot locomotion}.\hskip 1em plus
  0.5em minus 0.4em\relax CRC press, 2007.

\bibitem{da2017first}
X.~Da, R.~Hartley, and J.~W. Grizzle, ``First steps toward supervised learning
  for underactuated bipedal robot locomotion, with outdoor experiments on the
  wave field,'' in \emph{Proc. IEEE Int. Conf. Robot. Automat.}, 2017, pp.
  3476--3483.

\bibitem{nguyen2017dynamic}
Q.~Nguyen, A.~Agrawal, X.~Da, W.~C. Martin, H.~Geyer, J.~W. Grizzle, and
  K.~Sreenath, ``Dynamic walking on randomly-varying discrete terrain with
  one-step preview,'' in \emph{Robotics: Science and Systems}, 2017.

\bibitem{bloesch2013state}
M.~Bloesch, M.~Hutter, M.~A. Hoepflinger, S.~Leutenegger, C.~Gehring, C.~D.
  Remy, and R.~Siegwart, ``State estimation for legged robots-consistent fusion
  of leg kinematics and {IMU},'' in \emph{Robotics: Science and Systems},
  Berlin, Germany, June 2013.

\bibitem{fankhauser2014robot}
P.~Fankhauser, M.~Bloesch, C.~Gehring, M.~Hutter, and R.~Siegwart,
  ``Robot-centric elevation mapping with uncertainty estimates,'' in \emph{Int.
  Conf. Climbing and Walking Robots}, 2014, pp. 433--440.

\bibitem{bloesch2017state}
M.~A. Bloesch, ``State estimation for legged robots--kinematics, inertial
  sensing, and computer vision,'' Ph.D. dissertation, 2017.

\bibitem{fallon2014drift}
M.~F. Fallon, M.~Antone, N.~Roy, and S.~Teller, ``Drift-free humanoid state
  estimation fusing kinematic, inertial and lidar sensing,'' in \emph{IEEE-RAS
  Int. Conf. Humanoid Robots}.\hskip 1em plus 0.5em minus 0.4em\relax IEEE,
  2014, pp. 112--119.

\bibitem{nobiliheterogeneous}
S.~Nobili, M.~Camurri, V.~Barasuol, M.~Focchi, D.~G. Caldwell, C.~Semini, and
  M.~Fallon, ``Heterogeneous sensor fusion for accurate state estimation of
  dynamic legged robots,'' in \emph{Robotics: Science and Systems}, 2017.

\bibitem{roston1991dead}
G.~P. Roston and E.~P. Krotkov, ``Dead reckoning navigation for walking
  robots,'' Carnegie-Mellon University, Pittsburgh, PA, Robotics Institute,
  Tech. Rep., 1991.

\bibitem{anderson1979optimal}
B.~D. Anderson and J.~B. Moore, \emph{Optimal filtering}.\hskip 1em plus 0.5em
  minus 0.4em\relax Englewood Cliffs, 1979.

\bibitem{rotella2014state}
N.~Rotella, M.~Bloesch, L.~Righetti, and S.~Schaal, ``State estimation for a
  humanoid robot,'' in \emph{Proc. IEEE/RSJ Int. Conf. Intell. Robots
  Syst.}\hskip 1em plus 0.5em minus 0.4em\relax IEEE, 2014, pp. 952--958.

\bibitem{dissanayake2001solution}
G.~Dissanayake, P.~Newman, S.~Clark, H.~F. Durrant-Whyte, and M.~Csorba, ``A
  solution to the simultaneous localization and map building ({SLAM})
  problem,'' \emph{IEEE Trans. Robot. Automat.}, vol.~17, no.~3, pp. 229--241,
  2001.

\bibitem{eustice2006exactly}
R.~M. Eustice, H.~Singh, and J.~J. Leonard, ``Exactly sparse delayed-state
  filters for view-based {SLAM},'' \emph{IEEE Trans. Robot.}, vol.~22, no.~6,
  pp. 1100--1114, 2006.

\bibitem{forster2016manifold}
C.~Forster, L.~Carlone, F.~Dellaert, and D.~Scaramuzza, ``On-manifold
  preintegration for real-time visual--inertial odometry,'' \emph{IEEE Trans.
  Robot.}, vol.~33, no.~1, pp. 1--21, 2017.

\bibitem{durrant2006simultaneous}
H.~Durrant-Whyte and T.~Bailey, ``Simultaneous localization and mapping: part
  {I},'' \emph{IEEE Robot. Automat. Mag.}, vol.~13, no.~2, pp. 99--110, 2006.

\bibitem{cadena2016past}
C.~Cadena, L.~Carlone, H.~Carrillo, Y.~Latif, D.~Scaramuzza, J.~Neira, I.~Reid,
  and J.~J. Leonard, ``Past, present, and future of simultaneous localization
  and mapping: Toward the robust-perception age,'' \emph{IEEE Trans. Robot.},
  vol.~32, no.~6, pp. 1309--1332, 2016.

\bibitem{thrun2004simultaneous}
S.~Thrun, Y.~Liu, D.~Koller, A.~Y. Ng, Z.~Ghahramani, and H.~Durrant-Whyte,
  ``Simultaneous localization and mapping with sparse extended information
  filters,'' \emph{The Int. J. Robot. Res.}, vol.~23, no. 7-8, pp. 693--716,
  2004.

\bibitem{lupton2012visual}
T.~Lupton and S.~Sukkarieh, ``Visual-inertial-aided navigation for high-dynamic
  motion in built environments without initial conditions,'' \emph{IEEE Trans.
  Robot.}, vol.~28, no.~1, pp. 61--76, 2012.

\bibitem{lee2015introduction}
J.~M. Lee, \emph{Introduction to smooth manifolds}.\hskip 1em plus 0.5em minus
  0.4em\relax Springer, 2015.

\bibitem{chirikjian2011stochastic}
G.~S. Chirikjian, \emph{Stochastic Models, Information Theory, and Lie Groups,
  Volume 2: Analytic Methods and Modern Applications}.\hskip 1em plus 0.5em
  minus 0.4em\relax Springer Science \& Business Media, 2011.

\bibitem{absil2009optimization}
P.-A. Absil, R.~Mahony, and R.~Sepulchre, \emph{Optimization algorithms on
  matrix manifolds}.\hskip 1em plus 0.5em minus 0.4em\relax Princeton
  University Press, 2009.

\bibitem{barfoot2017state}
T.~D. Barfoot, \emph{State Estimation for Robotics}.\hskip 1em plus 0.5em minus
  0.4em\relax Cambridge University Press, 2017.

\bibitem{supplementary}
\BIBentryALTinterwordspacing
R.~Hartley, J.~Mangelson, L.~Gan, M.~Ghaffari~Jadidi, J.~M. Walls, R.~M.
  Eustice, and J.~W. Grizzle, ``Supplementary material: \\ legged robot
  state-estimation through combined kinematic and preintegrated contact
  factors,'' University of Michigan, Tech. Rep., Feb. 2017. [Online].
  Available:
  \url{https://arxiv.org/src/1712.05873/anc/supplementary-material-legged.pdf}
\BIBentrySTDinterwordspacing

\bibitem{dellaert2012factor}
F.~Dellaert, ``Factor graphs and {GTSAM}: A hands-on introduction,'' Georgia
  Institute of Technology, Tech. Rep., 2012.

\bibitem{kaess2012isam2}
M.~Kaess, H.~Johannsson, R.~Roberts, V.~Ila, J.~J. Leonard, and F.~Dellaert,
  ``{iSAM2}: Incremental smoothing and mapping using the {Bayes} tree,''
  \emph{The Int. J. Robot. Res.}, vol.~31, no.~2, pp. 216--235, 2012.

\bibitem{carlone2014eliminating}
L.~Carlone, Z.~Kira, C.~Beall, V.~Indelman, and F.~Dellaert, ``Eliminating
  conditionally independent sets in factor graphs: A unifying perspective based
  on smart factors,'' in \emph{Proc. IEEE Int. Conf. Robot. Automat.}\hskip 1em
  plus 0.5em minus 0.4em\relax IEEE, 2014, pp. 4290--4297.

\end{thebibliography}

\end{document}